\documentclass[10pt,twocolumn,letterpaper]{article}

\usepackage{iccv}
\usepackage{times}
\usepackage{epsfig}
\usepackage{graphicx}
\usepackage{amsmath,bm}
\usepackage{amssymb}
\usepackage{mathbbol}
\usepackage{multirow}
\usepackage{multicol} 
\usepackage{array}
\usepackage{makecell}
\usepackage[dvipsnames]{xcolor}
\usepackage{booktabs}
\usepackage{arydshln}
\usepackage{subcaption} 
\usepackage{diagbox}
\usepackage{pifont}%
\usepackage[accsupp]{axessibility}

\newcommand{\cmark}{\ding{51}}%
\newcommand{\xmark}{\ding{55}}%

\usepackage[pagebackref=true,breaklinks=true,letterpaper=true,colorlinks,bookmarks=false]{hyperref}

\iccvfinalcopy %

\def\iccvPaperID{2124} %

\ificcvfinal\fi

\begin{document}

\newcommand{\fracpartial}[2]{\frac{\partial #1}{\partial  #2}}
\newcommand{\norm}[1]{\left\lVert#1\right\rVert}
\newcommand{\innerproduct}[2]{\left\langle#1, #2\right\rangle}
\newcommand{\fan}[1]{\Vert #1 \Vert}
\newcommand{\qileft}{[\kern-0.15em[}
\newcommand{\qiLeft}{\left[\kern-0.4em\left[}
\newcommand{\qiright}{]\kern-0.15em]}
\newcommand{\qiRight}{\right]\kern-0.4em\right]}
\newcommand{\sign}{{\mbox{sign}}}
\newcommand{\diag}{{\mbox{diag}}}
\newcommand{\armin}{{\mbox{argmin}}}
\newcommand{\rank}{{\mbox{rank}}}
\renewcommand{\vec}{{\mbox{vec}}}
\newcommand{\st}{{\mbox{s.t.}}}
\newcommand{\<}{\left\langle}
\renewcommand{\>}{\right\rangle}
\newcommand{\lbar}{\left\|}
\newcommand{\rbar}{\right\|}
\renewcommand{\Roman}[1]{\uppercase\expandafter{\romannumeral#1}}
\newcommand{\red}[1]{{\color{red}{#1}}}
\newcommand{\blue}[1]{{\color{blue}{#1}}}

\renewcommand{\a}{{\bm{a}}}
\renewcommand{\b}{{\bm{b}}}
\renewcommand{\d}{{\bm{d}}}
\newcommand{\e}{{\bm{e}}}
\newcommand{\f}{{\bm{f}}}
\newcommand{\g}{{\bm{g}}}
\renewcommand{\o}{{\bm{o}}}
\newcommand{\p}{{\bm{p}}}
\newcommand{\q}{{\bm{q}}}
\renewcommand{\r}{{\bm{r}}}
\newcommand{\s}{{\bm{s}}}
\renewcommand{\t}{{\bm{t}}}
\renewcommand{\u}{{\bm{u}}}
\renewcommand{\v}{{\bm{v}}}
\newcommand{\w}{{\bm{w}}}
\newcommand{\x}{{\bm{x}}}
\newcommand{\y}{{\bm{y}}}
\newcommand{\z}{{\bm{z}}}
\newcommand{\balpha}{{\bm{\alpha}}}
\newcommand{\bbeta}{{\bm{\beta}}}
\newcommand{\bmu}{{\bm{\mu}}}
\newcommand{\bsigma}{{\bm{\sigma}}}
\newcommand{\blambda}{{\bm{\lambda}}}
\newcommand{\btheta}{{\bm{\theta}}}
\newcommand{\bgamma}{{\bm{\gamma}}}
\newcommand{\bxi}{{\bm{\xi}}}
\newcommand{\bphi}{{\bm{\phi}}}

\newcommand{\ba}{{\bm{A}}}
\newcommand{\bb}{{\bm{B}}}
\newcommand{\bc}{{\bm{C}}}
\newcommand{\bd}{{\bm{D}}}
\newcommand{\be}{{\bm{E}}}
\newcommand{\bg}{{\bm{G}}}
\newcommand{\bi}{{\bm{I}}}
\newcommand{\bj}{{\bm{J}}}
\newcommand{\bl}{{\bm{L}}}
\newcommand{\bo}{{\bm{O}}}
\newcommand{\bp}{{\bm{P}}}
\newcommand{\bq}{{\bm{Q}}}
\newcommand{\bs}{{\bm{S}}}
\newcommand{\bu}{{\bm{U}}}
\newcommand{\bv}{{\bm{V}}}
\newcommand{\bw}{{\bm{W}}}
\newcommand{\bx}{{\bm{X}}}
\newcommand{\by}{{\bm{Y}}}
\newcommand{\bz}{{\bm{Z}}}
\newcommand{\bTheta}{{\bm{\Theta}}}
\newcommand{\bSigma}{{\bm{\Sigma}}}

\newcommand{\A}{{\mathcal{A}}}
\newcommand{\B}{\mathcal{B}}
\newcommand{\C}{\mathcal{C}}
\newcommand{\D}{\mathcal{D}}
\newcommand{\F}{\mathcal{F}}
\renewcommand{\H}{\mathcal{H}}
\newcommand{\I}{\mathcal{I}}
\renewcommand{\L}{\mathcal{L}}
\newcommand{\N}{\mathcal{N}}
\renewcommand{\P}{\mathcal{P}}
\newcommand{\X}{\mathcal{X}}
\newcommand{\Y}{\mathcal{Y}}
\newcommand{\W}{\mathcal{W}}

\title{SimMatchV2: Semi-Supervised Learning with Graph Consistency}

\author{%
	Mingkai Zheng$^{1}$ \quad Shan You$^{2}$\thanks{Correspondence to: Shan You $<$\texttt{youshan@sensetime.com}$>$.} \\ \quad Lang Huang$^3$ \quad Chen Luo$^5$ \quad Fei Wang$^4$ \quad Chen Qian$^2$ \quad Chang Xu$^1$\\
	\normalsize $^1$School of Computer Science, Faculty of Engineering, The University of Sydney \\ 
	\normalsize $^2$SenseTime Research\quad
	\normalsize $^3$ The University of Tokyo  \\
	\normalsize $^4$University of Science and Technology of China \\
        \normalsize $^5$State Grid Anhui Electric Power Research Institute \\
}

\maketitle
\ificcvfinal\thispagestyle{empty}\fi

\begin{abstract}
   Semi-Supervised image classification is one of the most fundamental problem in computer vision, which significantly reduces the need for human labor. In this paper, we introduce a new semi-supervised learning algorithm - SimMatchV2, which formulates various consistency regularizations between labeled and unlabeled data from the graph perspective. In SimMatchV2, we regard the augmented view of a sample as a node, which consists of a label and its corresponding representation. Different nodes are connected with the edges, which are measured by the similarity of the node representations. Inspired by the message passing and node classification in graph theory, we propose four types of consistencies, namely 1) node-node consistency, 2) node-edge consistency, 3) edge-edge consistency, and 4) edge-node consistency. We also uncover that a simple feature normalization can reduce the gaps of the feature norm between different augmented views, significantly improving the performance of SimMatchV2. Our SimMatchV2 has been validated on multiple semi-supervised learning benchmarks. Notably, with ResNet-50 as our backbone and 300 epochs of training, SimMatchV2 achieves 71.9\% and 76.2\% Top-1 Accuracy with 1\% and 10\% labeled examples on ImageNet, which significantly outperforms the previous methods and achieves state-of-the-art performance. Code and pre-trained models are available at \href{https://github.com/mingkai-zheng/SimMatchV2}{https://github.com/mingkai-zheng/SimMatchV2}.
\end{abstract}

\section{Introduction}
In the past decades, deep learning has demonstrated its superior performance in various tasks \cite{imagenet_cvpr09, coco, pascal-voc-2007, yang2022deep}. However, its outstanding performance is usually based on the fully-supervised setting, which requires a large amount of data and annotations. In a real-world scenario, high-quality data annotations are fairly hard to collect since it is generally expensive, time-consuming, and sometimes require expert knowledge (~\eg medical images). Semi-supervised learning (SSL) has recently drawn much attention to the research community due to its efficiency in dealing with few labeled data and leveraging massive unlabeled data, significantly reducing the need for human labor. 

\begin{figure}[!t]
    \centering
    \includegraphics[width=1.0\linewidth]{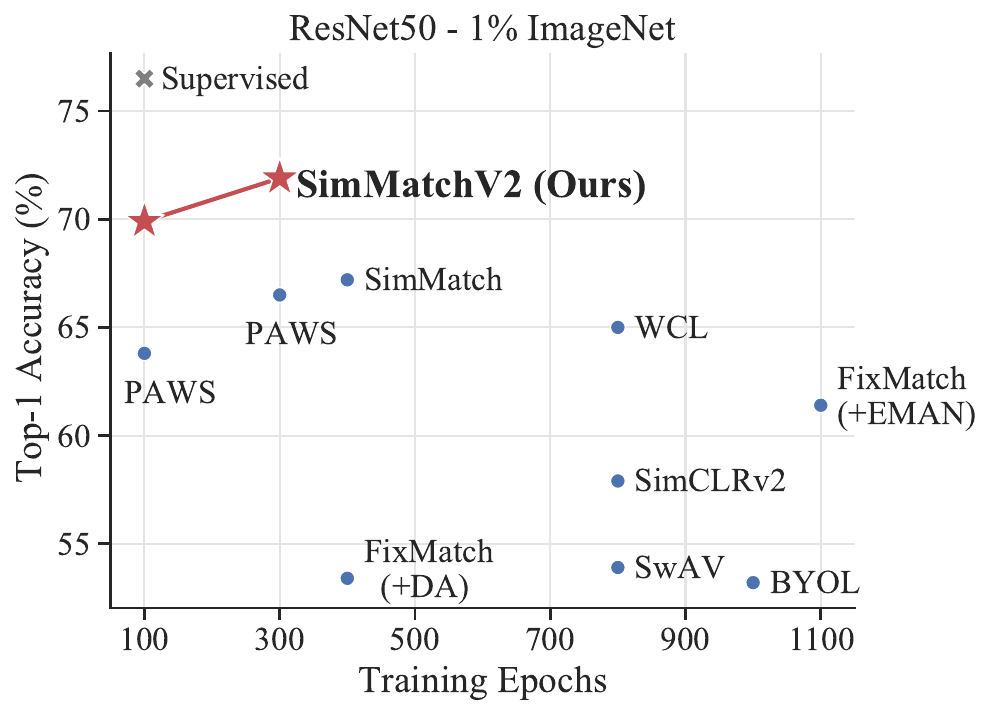}
    \vspace{-7mm}
    \caption{\footnotesize The Top-1 Accuracy of ResNet-50 on ImagetNet with 1\% labeled data. Our proposed SimMatchV2 performs significantly better than previous methods with fewer training epochs. Most importantly, 100 epochs of SimMatchV2 training only takes 14.7 GPU days, but the performance is much higher than PAWS \cite{paws} with 300 epochs (68 GPU days).}
    \vspace{-5mm}
    \label{fig:vs}
\end{figure}

For the typical SSL algorithms, one standard paradigm is to train a semantic classifier on the labeled data, then use the classifier to make a prediction on the unlabeled data as the pseudo-label \cite{pseudolabel}. Besides, a class-level ``consistency regularization" will be applied to enforce consistency between the class predictions of a model on different perturbations of the same input, which encourages the model to learn more robust and invariant features that generalize well to unseen data. Thus, the quality of pseudo-labels plays a crucial role in consistency regularization, and many works focus on generating stable and high-quality pseudo-labels. For example, MixMatch \cite{mixmatch}, ReMixMatch \cite{remixmatch}, and FixMatch \cite{fixmatch} are three typical  ``consistency regularization" based methods that significantly improve the performance of the SSL tasks. However, these methods highly depend on the classifier trained on the labeled data. A limited number of labels will result in a poor classifier that generates inaccurate pseudo-labels and leads to bad performance.

With the recent success of unsupervised visual representation learning \cite{simclr, moco, simclrv2, mocov2, byol, swav, SimSiam, greenmim, lewel, mae}, the pre-training and fine-tuning paradigm has gained popularity for the semi-supervised learning. In general, the model will be first trained on a large amount of unlabeled data with a proxy task (~\eg instance discrimination), which also imposes the consistency regularization on the instance level (\ie forcing the model to give similar representation for the different perturbations of the same instance). Then, a classifier will be attached to the pre-trained model, and the whole model will be fine-tuned with the limited labeled data. Many works \cite{mochi, PCL, adco, ressl, resslv2, DCL, alignment_uniformity, goodview, msf, prcl} have been proposed to design a better proxy task to improve the quality of the representation. Apparently, the biggest advantage of this approach is that it does not depend on the supervised trained classifier but still yields high-quality representations for unlabelled data. Nevertheless, the drawback of this approach is also very obvious, since no labels were used during the pre-training, making the model hard to converge. Generally, the pre-training requires more than 800 epochs to converge, which is extremely time-consuming.

Although many different types of consistency regularizations have been proposed, including class-level \cite{mixmatch, remixmatch, fixmatch}, instance-level \cite{simclr, moco, byol}, and those that consider both \cite{simmatch, fixmatch}, how to establish a complete and effective consistency regularizations remains unresolved.  This paper introduces a new SSL algorithm - SimMatchV2, which provides a complete consistency regularization solution for various scenarios. The overall idea is inspired by the message passing and node classification in the graph theory. In SimMatchV2, we consider the augmented view of a sample as a node in the graph which consists of a node label and its corresponding representation. Different nodes are connected with the edges, which are measured by the similarity of the node representations. Thus, the information of different nodes can be efficiently propagated through the graph. To this end, we propose four types of consistency regularization, namely 1) node-node consistency, 2) node-edge consistency, 3) edge-edge consistency, 
 and 4) edge-node consistency. Note that the 1) node-node consistency has been proposed in previous works, but our SimMatchV2 provides a complete and unified solution to design the consistencies. Moreover, we also propose a feature normalization technique to reduce the gaps between the feature norm between different augmented views and greatly improve the performance. Our contributions can be summarized as:
\begin{itemize}
 \vspace*{-0.3em}
 \setlength\itemsep{-0.1em}
 \item We propose SimMatchV2 - a novel semi-supervised algorithm that models various consistency regularizations from the graph perspectives and proposes four types of consistencies to provide a complete solution for the regularization in SSL.

 \item We show that a simple feature normalization reduces the gaps between the feature norm of different augmented views and greatly improves the performance.

 \item SimMatchV2 sets a new state-of-the-art performance on various SSL benchmarks. Notably, with ResNet-50 as our backbone and 300 epochs of training, SimMatchV2 achieves 71.9\% and 76.2\% Top-1 accuracy with 1\% and 10\% labeled data on ImageNet. This result is +4.7\% and +0.7\% better than prior arts.
 
\end{itemize}

\section{Related work}

\textbf{Consistency Regularization with Pseudo-Label.} Pseudo-label is a method that aims to generate labels for unlabeled data. As has been mentioned, it is often used in conjunction with consistency regularization, which constrains the model from making similar predictions on different perturbations of the same input. Most of these methods focus on generating high-quality pseudo-labels and exploring better perturbation strategies. For example, $\Pi$-model \cite{temporal} propose a temporal ensemble strategy for the pseudo-label to reduce the noise in the target. Mean-Teacher \cite{meanteacher} instead, maintain an exponential moving average (EMA) network to generate more reliable and stable pseudo-labels. Later, MixMatch \cite{mixmatch} was proposed and greatly improved the SSL performance. Specifically, it adopts the averaged prediction of multiple strongly augmented views of the same image as the pseudo-label, then incorporates with the Mixup \cite{mixup} augmentation that generates a convex combination of pairs of images and uses the corresponding mixed pseudo-label as the supervision target. ReMixMatch \cite{remixmatch} improves MixMatch by using only a weakly augmented view to generate the pseudo-label and also introduces the distribution alignment to encourage the marginal distribution of predictions on unlabeled data to be consistent with the labeled data. FixMatch \cite{fixmatch} abandons these complicated strategies and proposes a confidence-based threshold method to filter out the low-confidence pseudo-labels during the training.

\begin{figure*}[t]
    \centering
    \includegraphics[width=0.80\linewidth]{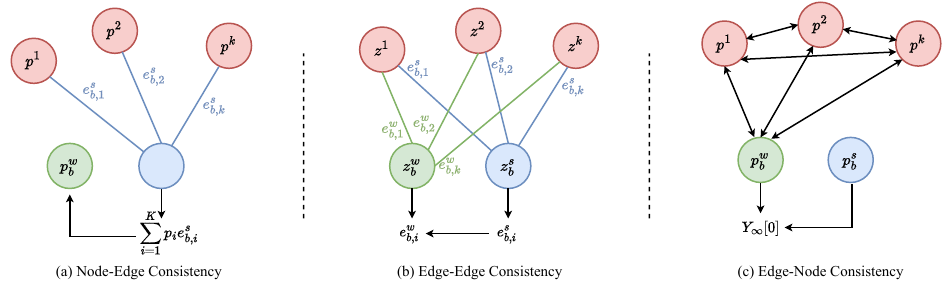}
    \vspace{-15pt}
    \caption{\footnotesize Illustration of various consistency in SimMatchV2. The green and blue nodes are from the weakly and strongly augmented view of the unlabeled data, the red nodes are the other samples.}
    \vspace{-14pt}
    \label{fig:SimMatchV2}
\end{figure*}

\textbf{Self-Supervised Learning for SSL.} Self-supervised learning has demonstrated its outstanding performance in learning visual representations without labels. It is intuitive and natural to leverage the proxy tasks from self-supervised learning to help with SSL tasks. As mentioned, the most straightforward solution is to pre-train the model in a self-supervised manner on a large amount of unlabeled data and then fine-tune the model with the limited labeled data. In this way, the design of the self-supervised task is particularly important. For example, instance discrimination \cite{instance_discrimination} is one of the most popular proxy tasks, which aims to encourage different augmented views of the same image to be pulled closer on embedding space but pushes apart all the other images away. BYOL \cite{byol} and SimSiam \cite{SimSiam} omit the negatives in contrastive learning and leverage an asymmetric structure to  avoid representation collapse. Recently, some works (\eg CoMatch \cite{comatch} and SimMatch \cite{simmatch}) have been proposed to jointly leverage both class-level and instance-level consistencies and greatly improve the SSL performance, especially in the case when only very limited labeled data is available. There are also many works \cite{eman, semivit} exploring the two-stage framework (\ie pre-training followed by fine-tuning with SSL algorithms like FixMatch), which shows the importance of the self-supervised pretrain to the SSL tasks.

\section{Methodology}

\subsection{Problem Formulation}
Similar to \cite{comatch, simmatch}, we define a batch of labeled data as $\mathcal{X}  = \{ x_b : b \in (1,...,B) \}$, and a batch of unlabeled data as $\mathcal{U}  = \{ u_b : b \in (1,...,\mu B) \}$, where $\mu$ is the coefficient of unlabeled batch size. Then, a neural network encoder will extract the feature $h$ from the augmented images. We use $h^{w}_{b}$ and $h^{s}_{b}$ to denote the feature extracted from the weakly and strongly augmented images, respectively. Then, we use $W \in $ to represent a linear layer followed by a SoftMax function that transforms the feature to a class probability \ie $p = \text{SoftMax}(W^{T}h)$. Finally, we adopt the $g(\cdot)$ to denote a two-layer non-linear projection head which maps the feature to a low-dimensional representation vector \ie $z = g(h)$. By following the common setting as in \cite{simclr, moco, simmatch, comatch}, the representation $z$ will be normalized to a unit vector by default. Note that the reason we use $z$ as the representation instead of $h$ is that we need to store a large number of samples as in \cite{moco}. The dimension of $z$ is much smaller than $h$ which makes the training more efficient. For example, $h$ is 2048-D and $z$ is 128-D for ResNet-50.

For labeled data, we perform standard supervised learning with the cross-entropy loss, which can be represented by Eq. \eqref{equation:loss_sup}, where $y_{b}$ is the one-hot ground truth label,
\begin{equation}
    \label{equation:loss_sup}
    \mathcal{L}_{s} = - \frac{1}{B} \sum_{b=1}^{B} \; y_{b} \log p_{b}^{w}.
\end{equation}

\subsection{Consistency Regularization with Graph}
SimMatchV2 considers the augmented view of a sample as a node in the graph, which consists of a node label $p$ (class prediction for unlabeled data and ground truth for labeled data) and the representation $z$. Different nodes are connected by edges measured by the similarity of the representations. We use strong/weak node to name the node from strongly/weakly augmented views.

Given a graph with $K$ nodes, we have the node labels $\{p_k: k \in (1,...,K) \}$ and the corresponding representations $\{z_k: k \in (1,...,K) \}$. In our implementation, the graph nodes will be stored in a memory bank as in \cite{moco, simmatch, comatch}.   We can store both labeled and unlabeled nodes. The node label will be the ground truth for labeled data and $p$ for unlabeled data. Suppose we have an unlabeled weak node with the representation $z^{w}_{b}$. We use $e^{w}_{b, i}$ to denote the edge between the $z^{w}_{b}$ and the $i^{th}$ node on the graph, which can be expressed by Eq. \eqref{equation:edge}. $t$ is a temperature parameter that controls the sharpness of the similarity distribution. We also use $e^{w}_{b}$ to denote edges between $z^{w}_{b}$ and all the nodes on the graph (\ie $e^{w}_{b, i}$ is a scalar and $e^{w}_{b} \in R^{1 \times K}$).
\begin{equation}
    \label{equation:edge}
    e^{w}_{b, i} = \frac{\exp(z^w_b \cdot z_{i} / t) }{\sum_{k=1}^{K}  \exp(z^w_b \cdot z_{k} / t) } 
\end{equation} 

\textbf{Node-Node Consistency.} \emph{(Weak Node vs. Strong Node.)} The node-node consistency aims to maintain the consistency of the label between the strong node and weak node of the same instance. Such consistency has been proposed in many previous works \cite{mixmatch, remixmatch, fixmatch}. In this work, we have renamed it and integrated it into our graph-based framework. Concretely, we follow the design in previous works; given unlabeled data,  the node-node consistency can be expressed by Eq. \eqref{equation:loss_nn}, where $\tau$ is the confidence threshold.
\begin{equation}
    \label{equation:loss_nn}
    \mathcal{L}_{nn} = - \frac{1}{\mu B} \sum_{b=1}^{\mu B} \mathbb{1} (\max p^w_{b} > \tau ) \; p^w_b \log p^s_b.
\end{equation}

\textbf{Node-Edge Consistency.} \emph{(Weak Node vs. Strong Edge.)} The purpose of the node-edge consistency is to construct an edge $e^{s}_{b}$ by using the representation $z^{s}_{b}$ from the strong node, then use the edge $e^{s}_{b}$ as the weight to aggregate the node label from the graph. We expect the aggregated node label can be consistent with the label $p^{w}_{b}$ from the weak node. Finally, the node-edge consistency will be constrained by the cross-entropy loss between $p_{b}^{w}$ and the aggregated label, which can be expressed by Eq. \eqref{equation:loss_ne}. 
\begin{equation}
    \label{equation:loss_ne}
    \mathcal{L}_{ne} = - \frac{1}{\mu B} \sum_{b=1}^{\mu B} \; p_{b}^{w} \log (\sum_{i=1}^{K}p_{i} \; e^{s}_{b, i} )
\end{equation}
Note that the aggregated label $\sum_{i=1}^{K}p_{i}e^{s}_{b, i}$ does not include the node label $p^{s}_{b}$ and $p^{w}_{b}$ from itself. The gradient will only be applied on $e^{s}_{b,i}$.

\textbf{Edge-Edge Consistency.} \emph{(Weak Edge vs. Strong Edge.)} The edge-edge consistency can be regarded as a simple extension of the node-node consistency. Basically, the node-node consistency aims to let the model predict invariant labels for the weak and strong nodes of the same instances. Similarly, the edges should also be invariant with respect to the strong and weak nodes of the same instance. Thus, the edge-edge consistency can be formulated as the cross-entropy loss between the similarity distribution from different augmented views, which can be written as Eq. \eqref{equation:loss_ee}.
\begin{equation}
    \label{equation:loss_ee}
    \mathcal{L}_{ee} = - \frac{1}{\mu B} \sum_{b=1}^{\mu B} \;  \sum_{i=1}^{K} e^{w}_{b,i} \log e^{s}_{b,i} 
\end{equation}
\textbf{Edge-Node Consistency.} \emph{(Weak Edge vs. Strong Node.)} This is somewhat similar to reversed operation to the node-edge consistency. Concretely, node-edge consistency requires the aggregated label $\sum_{i=1}^{K} p_{i} e^{s}_{b, i}$ from the strong node to be consistent with the label $p^{w}_{b}$ from the weak node. In contrast, edge-node consistency requires the label $p^{s}_{b}$ from the strong node to be consistent with the aggregated label $\sum_{i=1}^{K}p_{i}e^{w}_{b, i}$ from the weak node. Given an unlabeled weak node which has $z^{w}_{b}$ and $p^{w}_{b}$. Then, we concatenate the unlabeled node with the graph to get $Z=[z^{w}_{b}, z_{1}, ..., z_{k}]$, $Y=[p^{w}_{b}, p_{1}, ..., p_{k}]$. Next, we use the representation matrix $Z$ to obtain the pairwise edge matrix $A$ (a.k.a. affinity matrix) in the same way as Eq. \eqref{equation:edge}. Note that the difference between $e_{b}$ and $A$ is that $e_{b}$ only calculates the edges between the unlabeled data with respect to the node in the graph ($e^{w}_{b} \in R^{1 \times K}$) but $A$ also contain the edge information within the graphs. ($A \in R^{(1+K) \times (1+K)}$). We also want to emphasize that we will make $A$ a hollow matrix (\ie $A_{i,j} = 0$ if $ i == j$) before normalizing it to avoid the self-loop in the graph.  Here, we adopt the transductive inference \cite{transductive_inference} for aggregation:
\begin{equation}
    \label{equation:prop_onece}
\begin{aligned}
    Y_{\phi} &= \alpha \cdot A Y_{{\phi}-1} + (1 - \alpha) \cdot Y_{0} \\
    &=  (\alpha \cdot A)^{\phi-1} Y_{0} + (1-\alpha) \sum_{i=0}^{\phi-1}(\alpha \cdot A)^{i} Y_{0},
\end{aligned}
\end{equation} 
where $\phi$ is the number of iterations that the label has been propagated, $Y_{0}$ is the initial state $[p^{w}_{b}, p_{1}, ..., p_{k}]$, and $\alpha$ is a parameter in $(0, 1)$ controlling the weight of the aggregated label and the initial state $Y_{0}$. The node label will be propagated multiple times until it fully converges. When $\phi \rightarrow \infty$, we have $(\alpha \cdot A)^{\phi-1} = 0$ since $ 0 < \alpha < 1$. We can also make an approximation such that $\lim_{\phi \to \infty} \sum_{i=0}^{\phi-1}(\alpha \cdot A)^{i} = (I - \alpha \cdot A)^{-1}$, where $I$ is the identity matrix. Thus, the converged $Y_{\infty}$ can be derived as:
\begin{equation}
    \label{equation:prop_infty}
    Y_{\infty} = (1 - \alpha) (I - \alpha \cdot A)^{-1} Y_{0}
\end{equation}

Finally, the converged pseudo-label of the unlabeled data can be obtained from $Y_{\infty}[0]$ (\ie the first row vector of $Y_{\infty}$). Note that if we take $\alpha=1$, then $Y_{1}[0]$ is identical to $\sum_{i=1}^{K} p_{i} e^{w}_{b, i}$. Since  $Y_{\infty}[0]$ contains both information from the node label and aggregated label. Thus, we can directly use it to replace the original $p^{w}_{b}$ in Eq. \eqref{equation:loss_nn}. Moreover, we should notice an inverse operator in Eq. \eqref{equation:prop_infty}, which is computationally very expensive for a large matrix. Thus, to reduce the complexity, we will only take the Top-N nearest node from $Z$ and $Y$ instead of using the whole graph.

\textbf{Feature Normalization.} 
As we have presented in Eq. \eqref{equation:loss_nn}, one of the training objectives is to minimize the predictions between the weakly and strongly augmented views. However, significant gaps exist between the feature norms of different augmented views. Also, from $p = \text{SoftMax}(W^{T}h)$, we can observe that the norm of $h$ does not affect the prediction (~\ie $argmax (p)$). Thus, minimizing the gaps between the feature norm is unnecessary and brings additional complexity to the training. To address this issue, we simply apply a layer normalization \cite{layernorm} on the feature $h$, which can be expressed by Eq .\eqref{equation:layernorm}. Finally, the linear layer $W$ and non-linear projection head $g(\cdot)$ will be applied on top of the $\widehat{h}$,
\begin{equation}
    \label{equation:layernorm}
    \widehat{h} =  \text{LayerNorm}(h).
\end{equation}

Finally, the overall objective of SimMatchV2 can be expressed by Eq. \eqref{equation:loss_overall}, where $\lambda_{nn}$, $\lambda_{ee}$, and $\lambda_{ne}$ are the balancing factors that control the weights of the different losses.
\begin{equation}
    \label{equation:loss_overall}
    \mathcal{L} = \mathcal{L}_{s} + \lambda_{nn} \mathcal{L}_{nn} + \lambda_{ne} \mathcal{L}_{ne} + \lambda_{ee} \mathcal{L}_{ee}
\end{equation}

\renewcommand\arraystretch{0.7}
\begin{table*}
\centering
\caption{\footnotesize Error rate (\%) and Rank on the classical SSL benchmark. The error rate and standard deviation are reported based on three runs. The experimental settings in this table are all based on the USB \cite{usb} code base.   All results of other methods are directly copied from the USB \cite{usb} benchmark paper. The results for fully-supervised training on STL-10 is not provided since it contains an unlabeled set with unknown labels.
}
\vspace{-3mm}
\label{table:classic-cv-results}
\resizebox{\textwidth}{!}{
\begin{tabular}{l|ccc|ccc|ccc|ccc|c|c|c}
\toprule
Dataset & \multicolumn{3}{c|}{CIFAR-10}& \multicolumn{3}{c|}{CIFAR-100} & \multicolumn{3}{c|}{SVHN} & \multicolumn{3}{c|}{STL-10} & \multicolumn{1}{c|}{Friedman}   & \multicolumn{1}{c}{Final} & \multicolumn{1}{c}{Mean} \\ \cmidrule(r){1-1}\cmidrule(lr){2-4}\cmidrule(lr){5-7}\cmidrule(lr){8-10}\cmidrule(lr){11-13}\,

\# Label & \multicolumn{1}{c}{40} & \multicolumn{1}{c}{250} & \multicolumn{1}{c|}{4000}  & \multicolumn{1}{c}{400} & \multicolumn{1}{c}{2500} & \multicolumn{1}{c|}{10000} & \multicolumn{1}{c}{40} & \multicolumn{1}{c}{250}  & \multicolumn{1}{c|}{1000}  & \multicolumn{1}{c}{40}  & \multicolumn{1}{c}{250}   & \multicolumn{1}{c|}{1000} & \multicolumn{1}{c|}{rank}   & \multicolumn{1}{c}{rank} & \multicolumn{1}{c}{error rate} \\ \cmidrule(r){1-1}\cmidrule(lr){2-4}\cmidrule(lr){5-7}\cmidrule(lr){8-10}\cmidrule(l){11-13}\cmidrule(l){14-14}\cmidrule(l){15-15}\cmidrule(l){16-16}
Fully-Supervised &   \multicolumn{3}{c|}{4.57\tiny{±0.06}}                    & \multicolumn{3}{c|}{18.96\tiny{±0.06}}    & \multicolumn{3}{c|}{2.14\tiny{±0.01}}         & \multicolumn{3}{c|}{-}     &       -          &        -      & -        \\ 
Supervised       & 77.18\tiny{±1.32 }         & 56.24\tiny{±3.41}           & 16.10\tiny{±0.32}           & 89.60\tiny{±0.43}          & 58.33\tiny{±1.41}           & 36.83\tiny{±0.21}           & 82.68\tiny{±1.91}           & 24.17\tiny{±1.65}          & 12.19\tiny{±0.23}           & 75.4\tiny{±0.66}            & 55.07\tiny{±1.83}           & 35.42\tiny{±0.48}           &        -        &             -     & -    \\
\cmidrule(r){1-1}\cmidrule(lr){2-4}\cmidrule(lr){5-7}\cmidrule(lr){8-10}\cmidrule(l){11-13}\cmidrule(l){14-14}\cmidrule(l){15-15}\cmidrule(l){16-16}
$\Pi$-Model    \cite{temporal}    & 76.35\tiny{±1.69}         & 48.73\tiny{±1.07}         & 13.63\tiny{±0.07}         & 87.67\tiny{±0.79}          & 56.40\tiny{±0.69}          & 36.73\tiny{±0.05}          & 80.07\tiny{±1.22}         & 13.46\tiny{±0.61}         & 6.90\tiny{±0.22}          & 74.89\tiny{±0.57}         & 52.20\tiny{±2.11}         & 31.34\tiny{±0.64}         & 14.17  & 15 & 48.20  \\
Pseudo-Labeling \cite{pseudolabel} & 75.95\tiny{±1.86}         & 51.12\tiny{±2.91}         & 15.32\tiny{±0.35}         & 88.18\tiny{±0.89}          & 55.37\tiny{±0.48}          & 36.58\tiny{±0.12}          & 75.98\tiny{±5.36}         & 16.47\tiny{±0.59}         & 9.37\tiny{±0.42}          & 74.02\tiny{±0.47}          & 51.90\tiny{±1.87}         & 30.77\tiny{±0.04}         & 14.08  & 14 & 48.42  \\
Mean Teacher \cite{meanteacher}   & 72.42\tiny{±2.10}         & 37.56\tiny{±4.90}         & 8.29\tiny{±0.10}          & 79.96\tiny{±0.53}          & 44.37\tiny{±0.60}          & 31.39\tiny{±0.11}          & 49.34\tiny{±7.90}         & 3.44\tiny{±0.02}          & 3.28\tiny{±0.07}          & 72.90\tiny{±0.83}          & 49.30\tiny{±2.09}         & 27.92\tiny{±1.65}         & 11.75  & 12 & 40.01  \\
VAT         \cite{VAT}    & 78.58\tiny{±2.78}         & 28.87\tiny{±3.62}         & 10.90\tiny{±0.16 }        & 83.60\tiny{±4.21 }         & 46.20\tiny{±0.80}          & 32.14\tiny{±0.31}          & 84.12\tiny{±3.18}         & 3.38\tiny{±0.12}          & 2.87\tiny{±0.18}          & 73.33\tiny{±0.47}          & 57.78\tiny{±1.47}         & 40.98\tiny{±0.96}         & 12.92  & 13 & 45.23 \\
MixMatch      \cite{mixmatch}  & 35.18\tiny{±3.87}         & 13.00\tiny{±0.80}         & 6.55±\tiny{0.05}          & 64.91\tiny{±3.34}          & 39.29\tiny{±0.13}          & 27.74\tiny{±0.27}          & 27.77\tiny{±5.43}         & 4.69\tiny{±0.46}          & 3.85\tiny{±0.28}          & 49.84\tiny{±0.58}          & 32.05\tiny{±1.16}         & 20.17\tiny{±0.67}         & 11.00  & 11 & 27.09  \\
ReMixMatch   \cite{remixmatch}   & 8.13\tiny{±0.58}          & 6.34\tiny{±0.22}          & 4.65\tiny{±0.09}          & 41.60\tiny{±1.48}          & 25.72\tiny{±0.07}          & 20.04\tiny{±0.13}          & 16.43\tiny{±13.77}        & 5.65\tiny{±0.35}          & 5.36\tiny{±0.58}          & 27.87\tiny{±3.85}          & 11.14\tiny{±0.52}         & 6.44\tiny{±0.15}          & 7.75  & 10 & 14.95  \\
UDA       \cite{uda}      & 10.01\tiny{±3.34}         & 5.23\tiny{±0.08}          & 4.36\tiny{±0.09}          & 45.48\tiny{±0.37}          & 27.51\tiny{±0.28}          & 23.12\tiny{±0.45}          & 5.28\tiny{±4.02}          & \textbf{1.93\tiny{±0.03}} & \textbf{1.94\tiny{±0.02}} & 40.09\tiny{±4.03}          & 10.11\tiny{±1.15}         & 6.23\tiny{±0.28}         & 6.42  & 8 & 15.12  \\
FixMatch     \cite{mixmatch}   & 12.66\tiny{±4.49 }        & 4.95\tiny{±0.10 }         & 4.26\tiny{±0.01}          & 45.38\tiny{±2.07}          & 27.71\tiny{±0.42}          & 22.06\tiny{±0.10 }         & 3.37\tiny{±1.01}          & 1.97\tiny{±0.01}          & 2.02\tiny{±0.03 }         & 38.19\tiny{±4.76 }         & 8.64\tiny{±0.84 }         & 5.82\tiny{±0.06 }         & 5.29  & 5 & 14.75  \\
Dash       \cite{dash}     & 9.29\tiny{±3.28}          & 5.16\tiny{±0.28  }        & 4.36\tiny{±0.10  }        & 47.49\tiny{±1.05 }         & 27.47\tiny{±0.38}          & 21.89\tiny{±0.16}          & 5.26\tiny{±2.02 }         & 2.01\tiny{±0.01   }       & 2.08\tiny{±0.09}          & 42.00\tiny{±4.94     }     & 10.50\tiny{±1.37  }       & 6.30\tiny{±0.49}          & 6.83  & 9 & 15.32  \\
CoMatch    \cite{comatch}     & 6.51\tiny{±1.18}          & 5.35\tiny{±0.14  }        & 4.27\tiny{±0.12 }         & 53.41\tiny{±2.36  }        & 29.78\tiny{±0.11   }       & 22.11\tiny{±0.22  }        & 8.20\tiny{±5.32 }         & 2.16\tiny{±0.04 }         & 2.01\tiny{±0.04 }         & \textbf{13.74\tiny{±4.20}} & 7.63\tiny{±0.94} & 5.71\tiny{±0.08} & 5.75  & 6 & 13.41  \\
CRMatch     \cite{crmatch}    & 13.62\tiny{±2.62}         & \textbf{4.61\tiny{±0.17}} & \textbf{3.65\tiny{±0.04}} & 37.76\tiny{±1.45}          & \textbf{24.13\tiny{±0.16}} & \textbf{19.89\tiny{±0.23}} & \textbf{2.60\tiny{±0.77}} & 1.98\tiny{±0.04}          & 1.95\tiny{±0.03}          & 33.73\tiny{±1.17}          & 14.87\tiny{±5.09}         & 6.53\tiny{±0.36}          &4.08  &2 & 13.78  \\
FlexMatch    \cite{flexmatch}   & 5.29\tiny{±0.29 }         & 4.97\tiny{±0.07}          & 4.24\tiny{±0.06 }         & 40.73\tiny{±1.44 }         & 26.17\tiny{±0.18}          & 21.75\tiny{±0.15}          & 5.42\tiny{±2.83}          & 8.74\tiny{±3.32}          & 7.90\tiny{±0.30}          & 29.12\tiny{±5.04}          & 9.85\tiny{±1.35}          & 6.08\tiny{±0.34 }         &5.92  &7 & 14.19  \\
AdaMatch     \cite{adamatch}   & 5.09\tiny{±0.21} & 5.13\tiny{±0.05}          & 4.36\tiny{±0.05}          & 37.08\tiny{±1.35} & 26.66\tiny{±0.33}          & 21.99\tiny{±0.15}          & 6.14\tiny{±5.35}          & 2.13\tiny{±0.04}          & 2.02\tiny{±0.05}          & 19.95\tiny{±5.17}          & 8.59\tiny{±0.43}          & 6.01\tiny{±0.02}          &4.79  &3 & 12.18  \\
SimMatch    \cite{simmatch}    & 5.38\tiny{±0.01}          & 5.36\tiny{±0.08}          & 4.41\tiny{±0.07}          & 39.32\tiny{±0.72}          & 26.21\tiny{±0.37}          & 21.50\tiny{±0.11}          & 7.60\tiny{±2.11}          & 2.48\tiny{±0.61}          & 2.05\tiny{±0.05}          & 16.98\tiny{±4.24}          & 8.27\tiny{±0.40}          & 5.74\tiny{±0.31}          &5.25  &4 &12.10 \\
\textbf{SimMatchV2 (Ours)}       & \textbf{4.90\tiny{±0.16}}   & 5.04\tiny{±0.09}          & 4.33\tiny{±0.16}          & \textbf{36.68\tiny{±0.86}}          & 26.66\tiny{±0.38}          & 21.37\tiny{±0.20}          & 7.92\tiny{±2.80}          & 2.92\tiny{±0.81}          & 2.85\tiny{±0.91}          & 15.85\tiny{±2.62}          & \textbf{7.54\tiny{±0.81}}          & \textbf{5.65\tiny{±0.26}}          & \textbf{4.00}  & \textbf{1} & \textbf{11.81} \\
\bottomrule
\end{tabular}
}
\vspace{-4mm}
\end{table*}

\section{Experiments}

\subsection{Classical SSL Benchmark}

We will first conduct our experiments on the classic SSL benchmark which is widely adopted by most SSL papers. There are four datasets will be used in this experiment.

- CIFAR-10 \cite{cifar} consists of 60K 32x32 color images in 10 classes, with 6k images per class. The training and test set contains 5K and 1k samples, respectively. We conduct three different experiments on 40, 250, and 4,000 randomly selected images from the training set as the labeled data and use the rest of the training set as the unlabeled data.

- CIFAR-100 \cite{cifar} is a more complex version of CIFAR-10, which contains 100 classes with 600 images each. There are 500 training images and 100 testing images per class. The experimental setting are similar to CIFAR-10 above except using 400, 2,500, and 10K labeled data for training. 

- SVHN \cite{svhn} is a real-world image dataset that contains 32x32 cropped digits from house numbers in Google Street View images. It contains 73,257 training images, with 26,032 testing images and 531,131 extra images. We merge the training and extra set and randomly select 40, 250, and 1,000 from it as the label data to conduct the experiment.

- STL-10 \cite{stl10} consists of 5K labeled images, 100K unlabeled images, and 8,000 test images. We will randomly select 40, 250, and 1,000 from the 5K images; the rest images will be combined with 100K as the unlabeled data. 

\textbf{Implementation Details}. Our implementation is mainly based on the USB \cite{usb} code base for a fair comparison. Most of the hyper-parameters are inherited from \cite{fixmatch}. Specifically, we adopt WRN-28-2 \cite{wrn} for CIFAR-10 and SVHN, WRN-28-8 for CIFAR-100, and WRN-37-2 for STL-10. We use an SGD optimizer with Nesterov momentum \cite{sgdmomentum} for all experiments. The weight decay will be set to $5e^{-4}$ for most experiments except CIFAR-100, for which we set it as $1e^{-3}$. For the learning rate schedule, we use a cosine learning rate decay \cite{cosine_lr}, which adjusts the learning rate to $0.03 \cdot cos(\frac{7\pi s}{16S})$ where $s$ is the current training step, and $S$ is the total number of training steps. The model will be optimized for $2^{20}$ steps. We use 64 labeled data and 448 unlabeled data for each batch. All experiments will be conducted on a single GPU. The weak and strong augmentation policies are also directly inherited from the USB codebase. Note that the final performance will be reported based on the exponential moving average of the model with a 0.999 decay rate. For SimMatchV2 specific hyper-parameters, we set $\lambda_{nn}=1, \lambda_{ne}=1, \lambda_{ee}=1, t=0.1, \alpha=0.1, \tau=0.95, n=8$. ($n$ is the Top-N nearest node for Eq. \eqref{equation:prop_infty})  These hyper-parameters are kept unchanged for all experiments. We maintain two memory banks for storing both unlabeled nodes and labeled nodes. The unlabeled bank will be used for calculating the Eq. \eqref{equation:loss_ne} and Eq. \eqref{equation:loss_ee}. The labeled bank will be used for computing the Eq. \eqref{equation:prop_infty}.

\renewcommand\arraystretch{0.4}
\begin{table*}
    \footnotesize
	\centering
	\caption{\footnotesize Top-1 an Top-5 Accuracy with ResNet-50 on ImageNet with 1\% and 10\% settings. Note that the underline in the parameters - inference columns denotes the methods which require more parameters than standard ResNet-50 during the inference stage.}
        \vspace{-3mm}
	\begin{tabular}	{l | l | c | c | c | c | c  c | c  c }
	\toprule
	\multirow{3}{*}{\shortstack[c]{Pre-training\\Algorithm}} & \multirow{3}{*}{\shortstack[c]{Semi-supervised\\Algorithm}} & \multirow{3}{*}{\shortstack[c]{Pre-training\\Epochs}} & \multirow{3}{*}{\shortstack[c]{Semi-supervised\\Epochs}} & \multirow{3}{*}{\shortstack[c]{Paramters\\ train}} & \multirow{3}{*}{\shortstack[c]{Paramters\\inference}} & \multicolumn{2}{c|}{Top-1} & \multicolumn{2}{c}{Top-5} \\
	& & & & & & \multicolumn{2}{c|}{Label fraction} & \multicolumn{2}{c}{Label fraction} \\
	 & & & & & &  1\% & 10\%& 1\% & 10\%\\
	\midrule
        \multirow{8}{*}{None} & VAT+EntMin.~\cite{VAT,EntMin,S4L} & - & $\sim$100 & 25.6M & 25.6M & - & - & 47.0 & 83.4\\ 
        & S4L-Rotation~\cite{S4L} & - &  $\sim$200 & 25.6M & 25.6M & - & - & 53.4 & 83.8\\ 
	& UDA ~\cite{uda} & - & $\sim$530 &  25.6M & 25.6M & - & 68.8 & - & 88.8 \\
	& FixMatch ~\cite{fixmatch} & - & $\sim$300 &  25.6M & 25.6M & - & 71.5 & - & 89.1\\ 
        & FixMatch-EMAN ~\cite{eman} & - & $\sim$300 &  25.6M & 25.6M & - & 72.8 & - & 90.3\\ 
        & SsCL \cite{sscl} & - & $\sim$800  &  28.0M & 25.6M & 60.2 & 72.1 & 82.8 & 90.9 \\ 
	& CoMatch \cite{comatch} & - & $\sim$400  &  28.0M & 25.6M & 66.0 & 73.6 & 86.4 & 91.6 \\ 
    & SimMatch \cite{simmatch} & - & $\sim$400 & 28.0M & 25.6M & 67.2 & 74.4 & 87.1 & 91.7 \\
	\midrule
	PCL~\cite{PCL} & \multirow{6}{*}{Fine-tune} & $\sim$200  & $\sim$20 & 23.8M & 25.6M  & - & - & 75.3&85.6 \\ 
	SimCLR~\cite{simclr} & & $\sim$1000 & $\sim$60 &  28.0M & 25.6M  & 48.3& 65.6  & 75.5&87.8 \\ 
	SimCLR V2~\cite{simclrv2} & & $\sim$800 & $\sim$60 &  32.2M & \underline{29.8M}  & 57.9& 68.4  & 82.5 &89.2 \\ 
	BYOL~\cite{byol} & & $\sim$1000 & $\sim$50 & 35.1M & 25.6M  & 53.2&  68.8& 78.4&   89.0\\ 
	SwAV~\cite{swav} & & $\sim$800 & $\sim$20 & 28.0M & 25.6M  & 53.9& 70.2 &  78.5&   89.9\\
        WCL~\cite{WCL} & & $\sim$800 & $\sim$60 & 32.2M & \underline{29.8M}  & 65.0 & 72.0 &  86.3&   91.2\\\midrule
	\multirow{2}{*}{MoCo V2 ~\cite{mocov2}}  & Fine-tune & $\sim$800 & $\sim$20 &  28.0M & 25.6M  & 49.8 & 66.1& 77.2 & 87.9\\ 
	& FixMatch-EMAN \cite{eman} & $\sim$800 & $\sim$300 & 28.0M & 25.6M & 61.4 & 73.9 & 82.1 & 91.0\\ \midrule
    PAWS~\cite{swav} & Fine-tune & $\sim$300 & $\sim$50 & 36.1M & \underline{29.8M}  & 66.5 & 75.5 &  - & -\\ \midrule
    \multirow{2}{*}{None} & \multirow{2}{*}{\textbf{SimMatchV2 (Ours)} } & - & \textbf{$\sim$100} & \textbf{28.0M} & \textbf{25.6M} & \textbf{69.9} & \textbf{74.8} & \textbf{88.7} & \textbf{91.7} \\
    & & - & \textbf{$\sim$300} & \textbf{28.0M} & \textbf{25.6M} & \textbf{71.9} & \textbf{76.2} & \textbf{90.0} & \textbf{92.2} \\

	\bottomrule
    \end{tabular}
    \vspace{-4mm}
\label{table:imagenet_result}
\end{table*}

\textbf{Performance on Classical SSL Benchmark}. The overall results are reported in Table \ref{table:classic-cv-results}. We compare our SimMatchV2 against 14 SSL algorithms.  As in Table \ref{table:classic-cv-results}, it is very hard for an algorithm to get the best performance on all settings. Many algorithms already meet or exceed the supervised baseline in some settings (CIFAR-10 and SVHN); the performance difference would not be significant in this case. For a more rigorous and fair comparison, we follow the steps from USB \cite{usb}, which adopts the Friedman rank \cite{friedman1, friedman2} as the final evaluation metric. We also report the mean error rate for all settings. As can be seen, our SimMatchV2 achieves the best performance on 4 out of 12 settings and also has the highest Friedman rank and lowest mean error rate. CRMatch \cite{crmatch} also achieves a very good performance that ranked second in the final rank. It has the best performance on 5 out of the 12 settings, but it is less stable when the number of labeled data is limited, and the mean error rate is much worse than SimMatchV2. AdaMatch \cite{adamatch} and SimMatch \cite{simmatch} has a more stable performance under various conditions and they have a very close mean error rate compared to SimMatchV2. However, these two methods did not achieve any best results in this experiment, leading to a worse Friedman rank. Overall, the highest Friedman rank and lowest mean error rate for SimMatchV2 should demonstrate superiority in terms of both performance and stability.

\subsection{Experiments on ImageNet}
Next, we evaluate our SimMatchV2 on the large-scaled ImageNet dataset \cite{imagenet_cvpr09}, which contains 1,000 categories with 1.2M training images and 50K validation images. Concretely, we follow previous works \cite{comatch, simmatch, paws, eman} to take 1\% and 10\% samples from the training images as our labeled data and use the rest samples as the unlabeled data. To avoid the effects of different selections on the labeled data, we follow \cite{simclr, simclrv2, eman, paws, WCL} to adopt a pre-defined labeled / unlabeled splits \footnote{\url{https://www.tensorflow.org/datasets/catalog/imagenet2012_subset}} that has been adopted in most SSL papers.

\textbf{Implementation Details}. For ImageNet experiments, most of our hyper-parameter and settings are directly inherited from \cite{simmatch}. Specifically,  we use ResNet-50 \cite{resnet} as our default backbone and use the SGD with Nesterov momentum as the optimizer. We set the weight decay to 0.0001 during the training. For the learning rate, we warm up the model for 5 epochs until it reaches 0.03, and it will be decayed to 0 with the cosine scheduler \cite{cosine_lr}. By following \cite{eman, simmatch}, we use 64 labeled data and 320 unlabeled data in each batch; the data will be evenly distributed on 8 GPUs. For the hyper-parameter settings, we set $\lambda_{nn}=10, \lambda_{ne}=10, \lambda_{ee}=10,  t=0.1,  \alpha=0.1, \tau=0.7, n=128$ for $1\%$ setting, and we use $\lambda_{nn}=5, \lambda_{ne}=5$ for $10\%$ since we found the performance is slightly better. For strong augmentation, we adopt a multi-crop strategy as in \cite{resslv2}, which contains five crops with different sizes (224x224, 192x192, 160x160, 128x128, and 96x96); these crops will be augmented with the policy  in \cite{dino}. For weak augmentation, we follow the same strategy in \cite{simmatch, comatch}.

\textbf{Performance on ImageNet}. The performance has been shown in Table \ref{table:imagenet_result}. With 100 epochs of training, SimMatchV2 achieves 70.4\% and 74.8\% Top-1 accuracy with 1\% and 10\% labeled data, which already surpass the performance of CoMatch \cite{comatch}, and SimMatch \cite{simmatch} with 400 epochs. With 300 epochs of training, SimMatchV2 achieves 72.1\% and 76.2\% Top-1 accuracy, which improves absolute +5.2\% and +0.7\% over PAWS \cite{paws}. As we can observe that the improvement of 10\% setting is much lesser than 1\% setting. We doubt that this is because the performance of 76.2\% is already very close to the fully supervised setting (76.5\%), which should be the upper bound of the SSL tasks. Most importantly, our SimMatchV2 established a new state-of-the-art performance without any pre-training. 

\renewcommand\arraystretch{0.7}
\begin{table}[h]
    \centering
    \setlength\tabcolsep{5pt}
    \vspace{-1mm}
    \caption{\footnotesize Runtime Comparison. MC denotes whether we use the multi-crop strategy. The experiments are performed on 1\% ImageNet with 100 epochs of training. $\downarrow$ indicates lower is better and $\uparrow$ means higher is better.}
    \vspace{-3mm}
    \label{table:runtime}
    \footnotesize
    \begin{tabular}{c  c c c c}
        \toprule
        Method & MC & GPU Memory $\downarrow$ & GPU days $\downarrow$ & 1\% Top-1 $\uparrow$ \\ \midrule
        CoMatch \cite{comatch} & \xmark & 81G & 10.3 & 61.1  \\ 
        SimMatch \cite{simmatch} & \xmark & 40G & 7.7 & 61.2 \\ 
        \textbf{SimMatchV2}  & \xmark & \textbf{42G} & \textbf{7.8} & \textbf{63.7} \\ \hdashline
        PAWS \cite{paws} & \cmark & 1048G & 22.6 & 63.8 \\ 
        \textbf{SimMatchV2}  & \cmark & \textbf{83G} & \textbf{14.7} & \textbf{69.9} \\ \bottomrule
    \end{tabular}
    \vspace{-3mm}
\end{table}

\textbf{Efficiency Comparison}. Next, We also compare the training efficiency for different methods. The experiments are conducted with 8 Nvidia GPUs except for PAWS \cite{paws}, which requires at least 64 GPUs to run. We use mixed precision training for all methods. Note that CoMatch \cite{comatch} and SimMatch \cite{simmatch} have their own labeled / unlabeled splits, which makes the comparison a bit unfair. In this experiment, we adopt their officially released codebase to reproduce the performance on the same splits. 

We show the training statistics in Table \ref{table:runtime}. When we omit the multi-crops, the computational cost of SimMatchV2 is slightly higher compared to SimMatch \cite{simmatch} (\ie +2G memory, +0.1 GPU days), and it is much more efficient than CoMatch \cite{comatch}. We can also observe that SimMatchV2 has much better performance (+2.5\%) than CoMatch \cite{comatch} and SimMatch \cite{simmatch} even without the multi-crops. Moreover, when the multi-crops are included, the training time of SimMatchV2 is nearly doubled (7.8 days $\rightarrow$ 14.7 days), but it is still only 65\% of PAWS \cite{paws} (14.7 days / 22.6 days). The extra training cost of PAWS \cite{paws} is from the support set containing 960x7 labeled data per batch. Most importantly, our SimMatchV2 achieves similar performance with PAWS (63.7\% vs. 63.8\%) \cite{paws} with only 35\% training time (7.8 days / 22.6 days), and has an absolute improvement of +6.6\% with 65\% training time.

\renewcommand\arraystretch{0.7}
\begin{table}[h]
    \centering
    \setlength\tabcolsep{3pt}
    \vspace{-2mm}
    \caption{\footnotesize More experiments on different architectures. Note that we do not use any pre-training in this experiment.}
    \vspace{-3mm}
    \footnotesize
    \begin{tabular}{c c c c c c}
        \toprule
        Method & Arch & Params & Epochs & 1\% & 10\%  \\ \midrule
        Semiformer \cite{semiformer} & ViT-S + Conv & 40M & 600 & - & 75.5 \\ \midrule 
        ProbPseudo Mixup \cite{semivit} & ViT-Small & 22M & 100 & - & 70.9 \\ 
        \textbf{SimMatchV2} & \textbf{ViT-Small} & \textbf{22M} & \textbf{100} & \textbf{63.7} & \textbf{75.9} \\ \hdashline
        ProbPseudo Mixup \cite{semivit} & ViT-Base & 86M & 100 & - & 73.5 \\ 
        \textbf{SimMatchV2} & \textbf{ViT-Base} & \textbf{86M} & \textbf{100} & \textbf{65.3} & \textbf{76.8} \\ \hdashline
        ProbPseudo Mixup \cite{semivit} & ConvNext-T & 28M & 100 & - & 74.1 \\ 
        \textbf{SimMatchV2} & \textbf{ConvNext-T} & \textbf{28M} & \textbf{100} & \textbf{71.3} & \textbf{77.8} \\ \hdashline
        ProbPseudo Mixup \cite{semivit} & ConvNext-S & 50M & 100 & - & 75.1 \\ 
        \textbf{SimMatchV2} & \textbf{ConvNext-S} & \textbf{50M}  & \textbf{100} & \textbf{72.3} & \textbf{78.4} \\ \bottomrule
    \end{tabular}
    \vspace{-2mm}
    \label{table:different_architecture}
\end{table}

\textbf{Working with More Advanced Architectures}. With the recent successes in model architecture design, many more powerful models have been proposed, especially the vision transformer and its variant \cite{vit, deit, cait, convnext, swin}. Thus, we further apply our SimMatchV2 on ViT-Small \cite{deit}, ViT-Base \cite{vit}, ConvNext-T \cite{convnext}, and ConvNext-S \cite{convnext} to show the generality on different architectures. In this experiment, we follow \cite{semivit} to use 128 labeled data and 640 unlabeled per batch. We also adopt the AdamW \cite{AdamW} optimizer with the learning rate of $1.25e^{-4}$ and weight decay of $0.1$. We present the results in Table \ref{table:different_architecture}. As we can see,  SimMatchV2 with a better model can generally produce a better result. The exploration of cooperating with more advanced SSL pipeline \cite{semivit} and stronger data augmentations (\eg ProbPseudo Mixup \cite{semivit}) will be left as a future work since it requires more hyper-parameter turnings and more sophisticated designs, which is not the focus of this paper.

\begin{figure*}[t]
    \centering
    \subcaptionbox{\footnotesize Pseudo-Labels Accuracy}{\includegraphics[width=.28\linewidth]{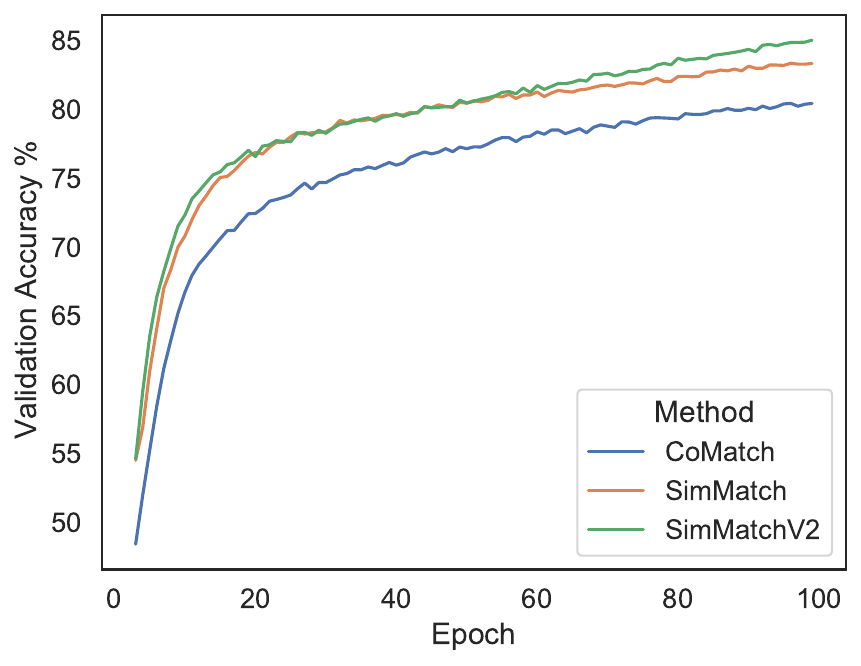}}
    \subcaptionbox{\footnotesize Unlabeled Data Accuracy}{\includegraphics[width=.28\linewidth]{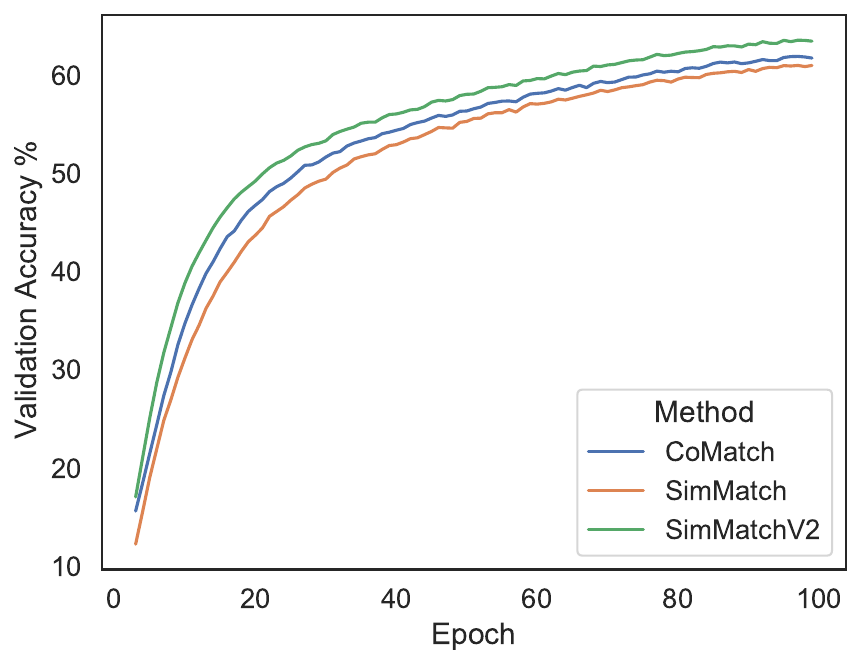}}
    \subcaptionbox{\footnotesize Validation Accuracy}{\includegraphics[width=.28\linewidth]{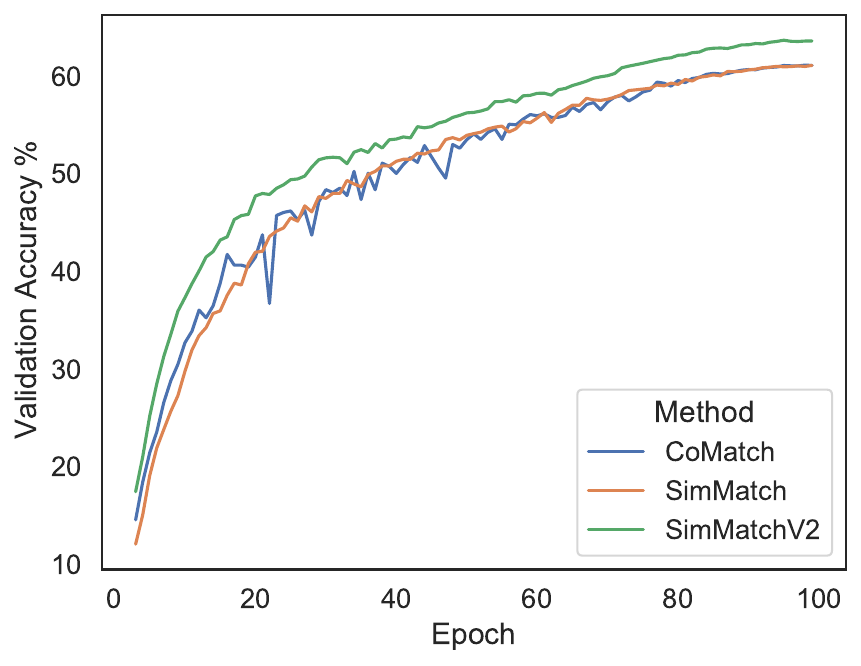}}
    \vspace{-3mm}
    \caption{\footnotesize In this Figure, we compare the training dynamics of SimMatchV2 with SimMatch \cite{simmatch}, and CoMatch \cite{comatch} from 3 different perspectives. a) The accuracy of $p^{w}$ that pass the threshold $\tau$. b) The accuracy of $p^{w}$ regardless of the threshold. c) The accuracy on the validation data. All experiments are in 1\% ImageNet setting with 100 epochs training. }
    \label{fig:training_dynamics}
    \vspace{-5mm}
\end{figure*}

\subsection{Experiments on Semi-iNat 2021}
\renewcommand\arraystretch{0.7}
\begin{table}[h]
\centering
\vspace{-5mm}
\caption{\footnotesize Accuracy for real-world dataset - Semi-iNat 2021.}
\vspace{-3mm}
\footnotesize
\begin{tabular}{ccccc}
\toprule
    \multirow{3}{*}{Method} & \multicolumn{4}{c}{Semi-iNat 2021} \\ 
    & \multicolumn{2}{c}{With Random Init} & \multicolumn{2}{c}{With MoCo Init} \\
\cline{2-5}
& Top1 & Top5 & Top1 & Top5 \\
\midrule
Supervised & 19.09 & 35.85 & 34.96 & 57.11 \\
\hline
MixMatch~\cite{mixmatch} & 16.89 & 30.83 & - & - \\
\hspace{0.23cm} +CCSSL~\cite{ccssl} & 19.65 & 35.09 & -  & - \\ \hdashline
FixMatch~\cite{fixmatch} & 21.41 & 37.65 & 40.30 & 60.05 \\
\hspace{0.23cm} +CCSSL~\cite{ccssl} & 31.21 & 52.25 & 41.28 & 64.30\\ \hdashline
CoMatch~\cite{comatch} & 20.94 & 38.96 & 38.94 & 61.85 \\
\hspace{0.23cm} +CCSSL~\cite{ccssl} & 24.12 & 43.23 & 39.85 & 63.68 \\ \hline
\textbf{SimMatchV2 (Ours)} & \textbf{43.05} & \textbf{63.81} & \textbf{53.13} & \textbf{73.40} \\
\bottomrule
\end{tabular}
\label{table:semi_inat}
\vspace{-1mm}
\end{table}

Finally, we also test the performance of SimMatchV2 on Semi-iNat 2021 \cite{semiinat}, which contains 810 different categories with 9,721 labeled data, 313,248 unlabeled data, and 4,050 validation data. Unlike the dataset we used in previous experiments, Semi-iNat 2021 is  designed to expose some challenges encountered in a real-world scenario, which contains highly similar classes, class imbalance, and domain mismatch between the labeled and unlabeled data. Such a more realistic dataset should be a better benchmark to reflect the performance of the different methods. 

In this experiment, most of the hyper-parameters and implementation details are the same as our ImageNet setting, except that we do not use the multi-crops strategy for fair comparison. The model will be optimized for $512 \times 1024$ steps as in \cite{ccssl}. We also follow the setting in \cite{ccssl} to conduct two experiments to report the training from scratch performance and training from pre-trained model performance.

We show the results in Table \ref{table:semi_inat}. Our SimMatchV2 achieves 43.05\% Top-1 accuracy when training from scratch, which is +11.84\% better than the state-of-the-art method (FixMatch + CCSSL). When the MoCo pre-trained model is adopted, our SimMatchV2 achieves a 53.13\% Top-1 accuracy, which is +10.08\% better than the from-scratch performance, and +11.85\% better than FixMatch + CCSSL. We believe this experiment will further demonstrate the superiority of SimMatchV2.

\section{Ablation study}
In this section, we will empirically study our SimMatchV2 based on various conditions and show the effect of each component and hyper-parameter sensitivities of our methods. For all experiments in this section, we adopt the ResNet-50 as our backbone and train the model on ImageNet for 100 epochs on the 1\% labeled data. The multi-crops are not used by default unless we mentioned.

\textbf{Training Dynamics}. We first show the training dynamics of SimMatchV2 compared to SimMatch \cite{simmatch} and CoMatch \cite{comatch}.  We visualize the pseudo-labels accuracy, unlabeled data accuracy, and validation accuracy in Figure \ref{fig:training_dynamics}. As we can observe that the pseudo-labels from SimMatchV2 have a higher quality than SimMatch \cite{simmatch}, especially in the late stage of the training. Both of these two methods generate better pseudo-labels than CoMatch \cite{comatch}. We can also notice that SimMatchV2 consistently has better accuracy for the unlabeled data and validation data, which shows the advantages of our method.

\renewcommand\arraystretch{0.7}
\begin{table}[h]
    \centering
    \vspace{-2mm}
    \caption{\footnotesize Effect of each component in SimMatchV2. N and E denote the node and edge, respectively. N-E means the node-edge consistency and vice versa. Feat norm means feature normalization. }
    \vspace{-3mm}
    \footnotesize
    \begin{tabular}{c c c c c c c}
        \toprule
        N-N & N-E & E-E & E-N & Feat Norm & Top-1 & Improvement  \\ \midrule
        \checkmark &   &   &   &   & 52.9  & - \\ 
        \checkmark & \checkmark  &   &   &   & 60.0 & +7.1       \\
        \checkmark & \checkmark & \checkmark  &   &   &  60.7 & +0.7       \\
        \checkmark & \checkmark  & \checkmark  & \checkmark  &   &  62.2 &  +1.5      \\
        \checkmark &  \checkmark & \checkmark  & \checkmark  & \checkmark  &  63.7 &   +1.5     \\
        \bottomrule
    \end{tabular}
    \vspace{-2mm}
    \label{table:components}
\end{table}

\textbf{Effect of Each Components}. In table \ref{table:components}, we show the effect of each design in SimMatchV2. Basically, if only node-node consistency is adopted, the setting is equivalent to the FixMatch \cite{fixmatch} with distribution alignment \cite{remixmatch}, which is our baseline in the first row of the table; Next, we can see that the node-edge, edge-node, and edge-node consistencies give +7.1\%, +0.7\%, +1.5\% improvements, respectively. Finally, the feature normalization also boosts the performance +1.5\%. This experiment should demonstrate that these different designs are all very effective for SSL tasks.

\renewcommand\arraystretch{0.8}
\begin{table}[h]
    \centering
    \setlength\tabcolsep{4.5pt}
    \vspace{-2mm}
    \caption{\footnotesize Effect of memory bank size on the performance. 5k - 163K are setting that we use unlabeled bank; labeled bank means we use all labeled bank (All labeled data 12K for 1\% ImageNet). }
    \vspace{-3mm}
    \footnotesize
    \begin{tabular}{c c c c c c c}
        \toprule
Bank Size  & 5K & 10K & 40K & 82K & 163K & Labeled Bank (12K)   \\ \midrule
Top-1       &  63.1 & 63.2 & 63.5 & \textbf{63.7} & 63.6 & 63.2   \\
        \bottomrule
    \end{tabular}
    \vspace{-2mm}
    \label{table:memory_bank}
\end{table}

\textbf{Node-Edge and Edge-Edge Consistency}. We will study the node-edge and edge-edge consistency together since the memory bank size is their primary factor. The experimental results are shown in Table \ref{table:memory_bank}. We can observe that when unlabeled data is used, the performance will be slightly increased with a larger memory bank, and the best memory bank size is 82K. We can also notice that we get slightly lower performance with 12K labeled data. Therefore, we decide to use the unlabeled data for node-edge and edge-edge consistency.

\renewcommand\arraystretch{0.8}
\begin{table}[h]
    \centering
    \setlength\tabcolsep{9.3pt}
    \vspace{-2mm}
    \caption{\footnotesize Effect of Top-N to Edge-Node Consistency. (We use labeled bank for this experiment.)}
    \vspace{-3mm}
    \footnotesize
    \begin{tabular}{c c c c c c c}
        \toprule
N           & 8 & 16 & 32 & 64 & 128 & 256   \\ \midrule
Top-1       & 63.4  & 63.5 & 63.5 & 63.5 & \textbf{63.7} & 63.6   \\
        \bottomrule
    \end{tabular}
    \vspace{-3mm}
    \label{table:topn}
\end{table}

\renewcommand\arraystretch{0.8}
\begin{table}[h]
    \centering
    \vspace{-2mm}
    \setlength\tabcolsep{4.5pt}
    \caption{\footnotesize Effect of labeled / unlabeled bank and $\phi$ in Eq. \eqref{equation:prop_onece} to Edge-Node Consistency.}
    \vspace{-3mm}
    \footnotesize
    \begin{tabular}{ c c c c c }
        \toprule
            & Labeled $Y_{\infty} $ & Labeled $Y_{1} $ & Unlabeled $Y_{\infty}$ & Unlabeled $Y_{1}$   \\ \midrule
    Top-1   & \textbf{63.7}    & 63.3  & 62.4   & 62.2             \\ \bottomrule
    \end{tabular}
    \vspace{-2mm}
    \label{table:label_unlabl_bank}
\end{table}

\textbf{Edge-Node Consistency}. As we have mentioned, it is computationally very expensive to involve the whole memory bank for the edge-node consistency calculation. Thus, we only adopt the Top-N nearest neighbor to propagate the label. We show this ablation study in Table \ref{table:topn}. The performance is very robust with the selection of the number of Top-N, the best results come from $N=128$, and the performance is only slightly decreased -0.3\% in the worst case. Note that all experiments in Table \ref{table:topn} are performed with the labeled bank; we also have the ablation study to test the effect of the labeled / unlabeled bank and the $\phi$ in Eq. \eqref{equation:prop_onece}. In Table \ref{table:label_unlabl_bank}, we can see that the performance with the labeled bank is much better than the unlabeled bank. Also, the performance with the fully converged $Y_{\infty}$ is +0.4\% better than $Y_{1}$, which shows the necessity of our design.

\renewcommand\arraystretch{0.8}
\begin{table}[h]
    \centering
    \vspace{-3mm}
    \caption{\footnotesize Effect of Feature Normalization}
    \vspace{-3mm}
    \footnotesize
    \begin{tabular}{c c c c }
        \toprule
       & Top-1  & Norm Gap $\downarrow$ & CE Loss $\downarrow$ \\ \midrule
W/o Feat Norm & 62.2  & 1.83  & 2.75  \\
W/\;\; Feat Norm  &  \textbf{63.7}   & \bf{1.31}  &  \bf{2.63}    \\
        \bottomrule
    \end{tabular}
    \vspace{-3mm}
    \label{table:feat_norm}
\end{table}

\textbf{Feature Normalization}. Finally, we show the effect of feature normalization in Table \ref{table:feat_norm}. As we mentioned, feature normalization aims to reduce the norm gaps between the weak and strong augmented views. Thus, to validate this, we calculate the mean of the absolute value of $\lVert h^{s} \rVert - \lVert h^{w} \rVert$ across the whole training process. We name it ``norm gaps" in Table \ref{table:feat_norm}. With the feature normalization, the norm gaps can be reduced from 1.83 $\rightarrow$ 1.31. We also calculate the mean of the CE loss Eq. \eqref{equation:loss_nn} (regardless of the threshold, \ie $\tau=0$) across the training process. 
\section{Conclusion}
In this work, we propose a new semi-supervised learning algorithm - SimMatchV2, which models consistency regularization from the graph perspective. SimMatchV2 provides four different types of consistency by leveraging the idea of message passing in graph theory and also proposes a simple feature normalization to reduce the gaps between the feature norm of different augmented views. The experiments on different benchmarks and various architectures demonstrate state-of-the-art performance. The limitation of this work is the missing exploration of cooperating with the more advanced SSL training pipelines and stronger data augmentations \cite{semivit} since it needs more experiments to find the optimal setting. We plan to investigate the pre-training setting in our future work.

\section*{Acknowledgment}
This work was supported in part by the Australian Research Council under Project DP210101859 and the University of Sydney Research Accelerator (SOAR) Prize.

{\small
\bibliographystyle{ieee_fullname}
\bibliography{egbib}
}

\end{document}